%% file: egpaper_final.tex
\documentclass[10pt,twocolumn,letterpaper]{article}

\usepackage{iccv}
\usepackage{times}
\usepackage{epsfig}
\usepackage{graphicx}
\usepackage{amsmath}
\usepackage{amssymb}

\usepackage{kotex}
\usepackage{booktabs}
\usepackage{multirow}
\usepackage{caption}
\usepackage{subcaption}

\usepackage[breaklinks=true,bookmarks=false]{hyperref}

\usepackage[capitalize]{cleveref}
\crefname{section}{Sec.}{Secs.}
\Crefname{section}{Section}{Sections}
\Crefname{table}{Table}{Tables}
\crefname{table}{Tab.}{Tabs.}

\input{macro}

\iccvfinalcopy % *** Uncomment this line for the final submission

\def\iccvPaperID{10867} % *** Enter the ICCV Paper ID here

\ificcvfinal\fi

\begin{document}

%%%%%%%%% TITLE
\title{3D-Aware Generative Model for Improved Side-View Image Synthesis}

\author{Kyungmin Jo$^{1,\ast}$ \and Wonjoon Jin$^{2,\ast}$ \and Jaegul Choo$^{1}$ \and Hyunjoon Lee$^{3}$ \and Sunghyun Cho$^{2}$\\
\and
$^1$KAIST\\
Daejeon, Korea\\
{\tt\small \{bttkm, jchoo\}@kaist.ac.kr}
\and
$^{2}$POSTECH\\
Pohang, Gyeongbuk, Korea\\
{\tt\small \{jinwj1996, s.cho\}@postech.ac.kr}
\and
$^{3}$Kakao Brain\\
Seongnam-si, Gyeonggi-do, Korea\\
{\tt\small malfo.lee@kakaobrain.com}
}

% Teaser
\twocolumn[{%
\renewcommand\twocolumn[1][]{#1}%
\maketitle
\begin{center}
    \centering
    \captionsetup{type=figure}
    \includegraphics[width=\textwidth]{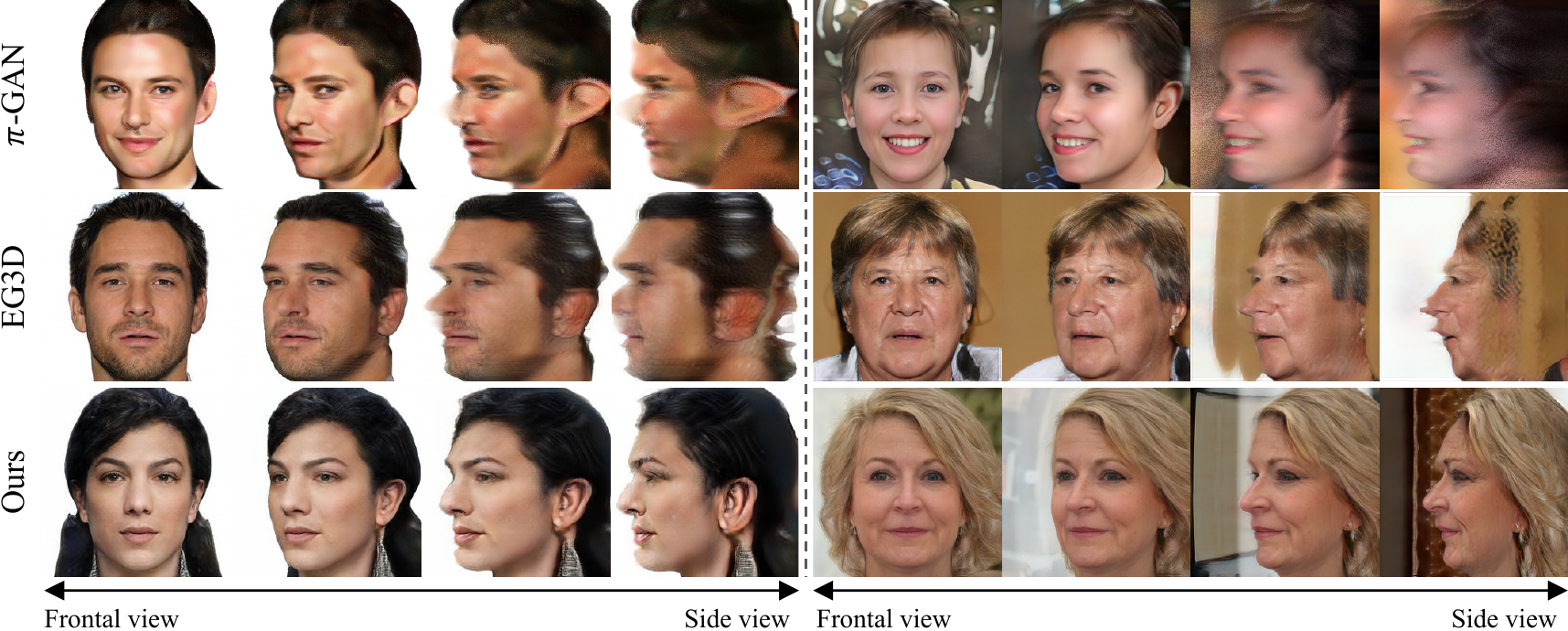}
    \caption{
    Our method robustly produces high-quality images of human faces, regardless of the camera pose while the baselines ($\pi$-GAN~\cite{chan2021pi} and EG3D~\cite{chan2022efficient}) generate blurry images at the steep pose.
    The images are rendered with horizontal rotation from the frontal view to the side view.
    }
    \label{fig:teaser}
\end{center}%
}]

\maketitle
\ificcvfinal\thispagestyle{empty}\fi
{\let\thefootnote\relax\footnotetext{\noindent ${}^{\ast}$Both authors contributed equally to this research. Also, this work was done during an internship at Kakao Brain.}}
%%%%%%%%% ABSTRACT
\begin{abstract}
    While recent 3D-aware generative models have shown photo-realistic image synthesis with multi-view consistency, the synthesized image quality degrades depending on the camera pose (e.g., a face with a blurry and noisy boundary at a side viewpoint). Such degradation is mainly caused by the difficulty of learning both pose consistency and photo-realism simultaneously from a dataset with heavily imbalanced poses. In this paper, we propose SideGAN, a novel 3D GAN training method to generate photo-realistic images irrespective of the camera pose, especially for faces of side-view angles. To ease the challenging problem of learning photo-realistic and pose-consistent image synthesis, we split the problem into two subproblems, each of which can be solved more easily. Specifically, we formulate the problem as a combination of two simple discrimination problems, one of which learns to discriminate whether a synthesized image looks real or not, and the other learns to discriminate whether a synthesized image agrees with the camera pose. Based on this, we propose a dual-branched discriminator with two discrimination branches. We also propose a pose-matching loss to learn the pose consistency of 3D GANs. In addition, we present a pose sampling strategy to increase learning opportunities for steep angles in a pose-imbalanced dataset. With extensive validation, we demonstrate that our approach enables 3D GANs to generate high-quality geometries and photo-realistic images irrespective of the camera pose.
\end{abstract}
\vspace{-5mm}

%%%%%%%%% BODY TEXT
\input{main_tex/1.introduction}
\input{main_tex/2.related_work}

\input{main_tex/3.method_framework}
\input{main_tex/4.method_training}
\input{main_tex/5.experiments}
\input{main_tex/6.conclusion}

\paragraph{
Acknowledgement}
{
This work was supported by the National Research Foundation of Korea (NRF) grant (NRF-2018R1A5A1060031), the Institute of Information \& communications Technology Planning \& Evaluation (IITP) grant (No.2019-0-01906, Artificial Intelligence Graduate School Program(POSTECH)) funded by the Korea government (MSIT), Institute of Information \& communications Technology Planning \& Evaluation (IITP) grant funded by the Korea government (MSIT) (No.2019-0-00075, Artificial Intelligence Graduate School Program (KAIST)), and the National Research Foundation of Korea (NRF) grant funded by the Korea government (MSIT) (No. NRF-2022R1A2B5B02001913).
}

{\small
\bibliographystyle{ieee_fullname}
\bibliography{egbib}
}

\end{document}

%% file: macro.tex
\def\MethodName{SideGAN}

\usepackage{xcolor}

%% file: main_tex/1.introduction.tex
\section{Introduction}
\label{sec:intro}

% 1. Explanation of GAN and multi-view consistent image generation
Generative Adversarial Networks (GANs)~\cite{goodfellow2020generative} have shown remarkable success in photo-realistic image generation~\cite{karras2019style, Karras2019stylegan2} by learning the distributions of high-resolution image datasets.
Recent studies have taken this success one step further by extending GANs to pose-controllable
image generation based on the guidance of a 3DMM prior~\cite{tewari2020stylerig, deng2020disentangled} or a differentiable renderer~\cite{zhang2020image}.
However, they produce inconsistent results across different poses and also suffer from limited pose controllability as they learn to generate 2D images for different poses independently without considering the 3D face structure.

%---------------------------------------------------------------------------------
% 2. Advent and trends of 3D GAN 
Therefore, 3D-aware GANs have emerged to achieve multi-view consistent image generation. Recent studies~\cite{schwarz2020graf, chan2021pi, gu2021stylenerf, xu20223d, chan2022efficient, skorokhodov2022epigraf, niemeyer2021giraffe} have tackled this problem by modeling the 3D structure of a face using neural radiance fields~\cite{mildenhall2021nerf}, enabling explicit view control. 
Combining volumetric feature projection with convolutional neural networks (CNNs) enables 3D GANs to generate photo-realistic face images in high resolution~\cite{gu2021stylenerf, or2022stylesdf, chan2022efficient}. Albeit their ability to synthesize photo-realistic images with explicit view control, their results do not have a stable quality depending on the camera pose (\cref{fig:teaser}). 
To be specific, side-view facial images generated by such methods show degraded qualities compared to photo-realistic images of frontal viewpoints ({\it e.g.,} a blurry and a noisy facial boundary).

\begin{figure}[!t]
\includegraphics[width=0.95\linewidth]{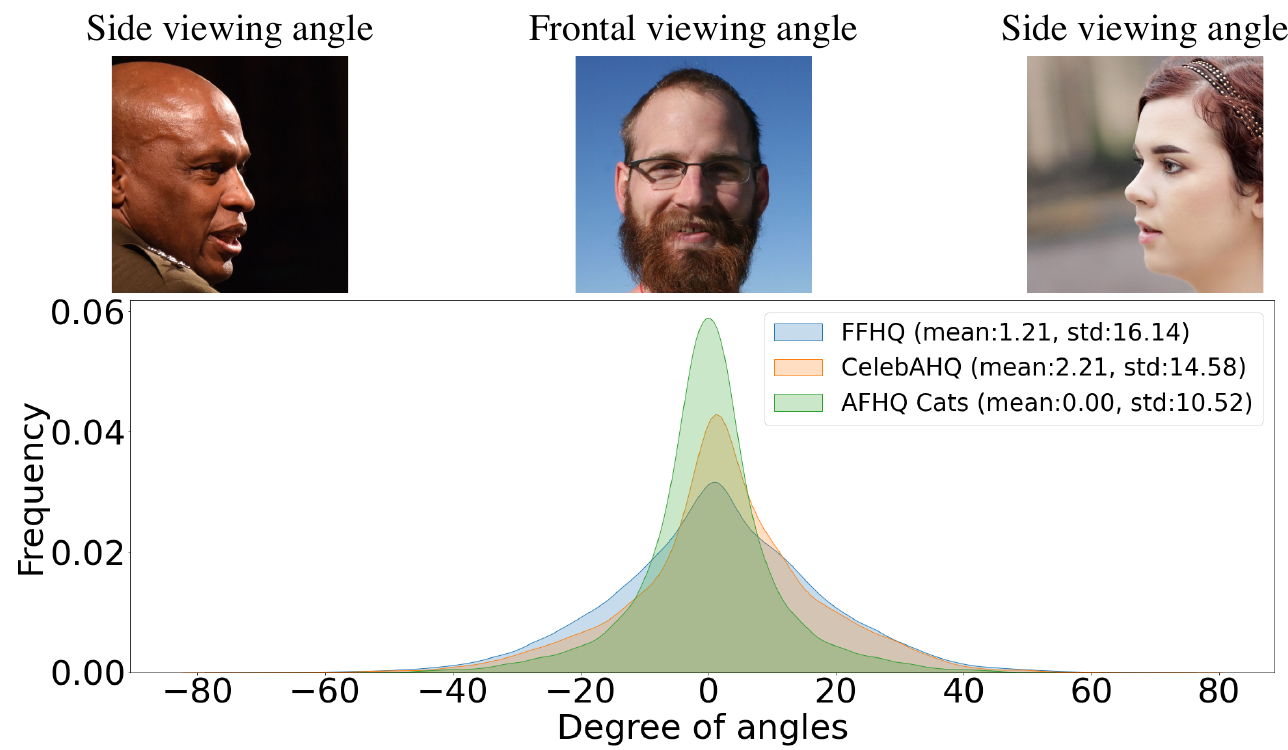}
% \vspace{-0.2cm}
\caption{Real-world face datasets generally have an imbalanced pose distribution, which is mainly concentrated on the frontal viewpoint.
}
\label{fig:pose_dist}
\vspace{-5mm}
\end{figure}

%---------------------------------------------------------------------------------
% 3. Our main task and tackling pointsds to ignore not enough samples in the datasets.
This unstable image quality is caused by the challenge for 3D-aware GANs to simultaneously learn to generate {\it pose-consistent} and {\it photo-realistic} images from a pose-imbalanced dataset (\cref{fig:pose_dist}) such as the FFHQ dataset~\cite{karras2019style} where most images are frontal-view images.
Specifically, EG3D~\cite{chan2022efficient}, the state-of-the-art 3D GAN approach, formulates the problem as a learning problem of a pose-conditional distribution of real images.
Unfortunately, learning the distribution of real images for each pose can be extremely challenging, especially for poses with only a small number of real images.
GRAM~\cite{deng2022gram} casts the problem as a combination of the learning of real/fake image discrimination and pose estimation.
Nevertheless, pose estimation from degraded side-view images is not trivial to learn either.
As a result, images generated by the existing 3D GANs are blurry or have noisy boundaries in the face region at steep pose angles (\cref{fig:teaser}).

To tackle this problem, we propose \MethodName{}, a novel 3D GAN training method to generate photo-realistic images irrespective of the viewing angle.
Our key idea is as follows.
To ease the challenging problem of learning photo-realistic and multi-view consistent image synthesis, we split the problem into two subproblems, each of which can be solved more easily.
Specifically, we formulate the problem as a combination of two simple discrimination problems, one of which learns to discriminate whether a synthesized image looks real or not, and the other learns to discriminate whether a synthesized image agrees with the camera pose.
Unlike the formulations of the previous methods, which try to learn the real image distribution for each pose, or to learn pose estimation, our subproblems are much easier as each of them is analogous to a basic binary classification problem.

Based on this key idea, we propose a {\it dual-branched discriminator}, which has two branches for learning photo-realism and pose consistency, respectively.
As these branches are supervised explicitly for their respective purposes, high-quality images with pose consistency can be produced at each viewing angle, and consequently, the generator creates high-quality images and shapes.
In addition, we propose a {\it pose-matching loss} to give supervision to the discriminator for the pose consistency, by considering a positive pose ({\it i.e.}, rendering pose or ground truth pose) and a negative pose ({\it i.e.}, irrelevant pose) for a given image.
For example, the frontal viewpoint is one of the irrelevant poses for a side-view image.
As reported in the experiments, this loss helps improve image and shape quality.
Compared to the previous pose estimation strategy~\cite{deng2022gram}, our pose-matching loss provides a more effective way to learn pose-consistent image generation, as the pose-matching loss casts the learning of pose-consistent image generation as the learning of simple binary classification that is much easier than the learning of accurate pose regression.

Additionally, we suggest a simple but effective training strategy to alleviate the degradation caused by insufficient semantic knowledge at steep poses in a pose-imbalanced dataset. As shown in \cref{fig:pose_dist}, most in-the-wild face datasets~\cite{karras2019style, karras2017progressive, choi2020stargan} usually have pose distributions concentrated on the frontal angle, causing the degradation of generated images at steep poses.
While we may construct a pose-balanced dataset in a controlled environment, it requires a significant amount of effort, and is also hard to guarantee the diversity like in the in-the-wild datasets.
Instead, we present an additional uniform pose sampling (AUPS) strategy that draws camera poses from both a uniform distribution and the actual camera pose distribution to enhance learning opportunities for steep angles during training.
Our experiments show that this simple pose sampling strategy substantially improves the generation quality for side-view images.

%---------------------------------------------------------------------------------

% 5. Additional benefits, Itemized Contributions.
Our contributions are summarized as follows:
\vspace{-1mm}
\begin{itemize}
    \setlength\itemsep{-0.3em}
    \item We split the problem of learning of 3D GANs into two easier subproblems: real/fake image discrimination and pose-consistency discrimination.
    \item We propose a dual-branched discriminator and a pose-matching loss to effectively learn the pose consistency by considering both positive and negative poses of a given image.
    \item We also present a simple but effective pose sampling strategy to compensate for the insufficient amount of side-view images in pose-imbalanced in-the-wild datasets.
    \item With extensive evaluations, \MethodName{} shows the state-of-the-art image and shape quality irrespective of the camera pose, especially at steep view angles.    
\end{itemize}

%% file: main_tex/2.related_work.tex
\section{Related work}
\label{sec:related}
\noindent{\bf Extending 2D GANs to have pose controllability.}
GANs~\cite{goodfellow2020generative} have achieved significant success in photo-realistic 2D image generation~\cite{karras2019style,Karras2019stylegan2}.
Extending 2D GANs to provide pose controllability
% multi-view consistent image generation 
has been addressed by disentangling 3D information from GAN's latent space.
Finding meaningful directions for editing pose in the latent space can be done with supervision from pre-trained classifiers~\cite{shen2020interpreting} or in an unsupervised manner~\cite{shen2021closed}. 
Editing the camera pose can be implemented by disentangling the pose factor from the latent space with guidance from a 3DMM prior~\cite{tewari2020stylerig, deng2020disentangled}. 
Zhang et al.~\cite{zhang2020image} utilize inverse graphics with a differentiable renderer for pose-controllable image generation by fine-tuning StyleGAN to have disentangled pose attributes.
Shi et al.~\cite{shi20223d_reb} exploit a depth prior to disentangle the latent codes of geometry and appearance for RGBD generation with pose controllability.
Unfortunately, these studies based on 2D GANs fundamentally lack multi-view consistency or accurate pose controllability since they do not consider the 3D structure of faces. 

\noindent{\bf 3D-aware GANs.}
Recent work incorporating neural 3D representations into GANs enables multi-view consistent image generation with explicit camera control.
GRAF~\cite{schwarz2020graf} and $\pi$-GAN~\cite{chan2021pi} adopt fully implicit volumetric fields with differentiable volumetric rendering for 3D scene generation.
However, these methods suffer from a large memory burden due to fully implicit networks, restricting image resolution and expressiveness.
To enable high-resolution image synthesis, GRAM~\cite{deng2022gram} restricts point sampling to regions near the learned implicit surface.
StyleNeRF~\cite{gu2021stylenerf}, StyleSDF~\cite{or2022stylesdf} and GIRAFFE~\cite{niemeyer2021giraffe} combine CNN-based upsamplers with volumetric feature projection in their multi-view consistent image generation.
EG3D~\cite{chan2022efficient}, which is the most recent and related to our work, achieves photo-realistic image synthesis based on their tri-plane representation and StyleGAN feature generator. 
While previous 3D GAN studies have made significant progress in 3D-aware image synthesis, they have a limitation that the image quality degrades as the viewpoint shifts from frontal angles to steeper angles. % in pose-imbalanced distribution.
To the best of our knowledge, our work is the first one to tackle the ineffectiveness of training 3D GANs from a pose-imbalanced dataset for photo-realistic multi-view consistent image generation irrespective of the camera pose.

%% file: main_tex/3.method_framework.tex
\section{\MethodName{} Framework}
\label{sec:method}
\begin{figure*}[t!]
\centering
%\begin{tabular}{@{}c}
\includegraphics[width=0.85\linewidth]{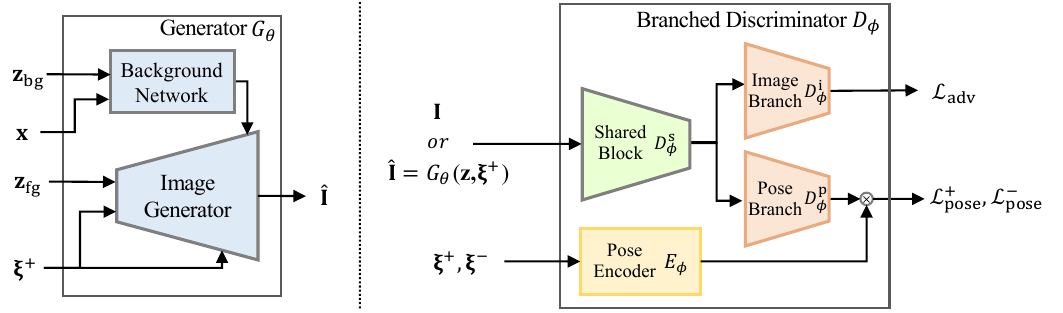}
%\end{tabular}
\vspace{-3mm}
\caption{Illustration of main architecture. 
The generator takes latent codes $\mathbf{z}_{fg}$ and $\mathbf{z}_{bg}$, camera parameters $\boldsymbol{\xi}^+$, and 3D position $\mathbf{x}$ as inputs and synthesizes an image $\hat{\mathbf{I}}$. The dual-branched discriminator takes either a real image $\mathbf{I}$ or a generated image $\hat{\mathbf{I}}$ and camera parameters $\boldsymbol{\xi}$ and outputs separably logits for image distribution and image-pose consistency.
}
\label{fig:main_architecture}
%\vspace{-5mm}
\end{figure*}

%%%%%%%%%%%%%%%%%%%%%%%%%%%
% overview
%%%%%%%%%%%%%%%%%%%%%%%%%%%
Our framework generates photo-realistic images irrespective of the camera pose even though most images in the training dataset are frontal-view images.
As shown in \cref{fig:main_architecture}, the main architecture is composed of two components. The first component is a generator $G_\theta$ for generating images from latent vectors $\mathbf{z}_{fg}$ and $\mathbf{z}_{bg}$ for the foreground and background regions, respectively, and a rendering camera parameter $\boldsymbol{\xi}^{+}$. The second component is a dual-branched discriminator $D_\phi$ for discriminating a generated image $\hat{\mathbf I}$ from a real image ${\mathbf I}$ and for discriminating whether a generated image agrees with a camera pose $\boldsymbol{\xi}$.
In the following, we describe each component.
More details on our framework are provided in Sec.~B.

%%%%%%%%%%%%%%%%%%%%%%%%%%%
% Generator
%%%%%%%%%%%%%%%%%%%%%%%%%%%
\subsection{Generator}

Existing 3D GAN models~\cite{chan2022efficient,chan2021pi} mainly render the background and foreground together by a single network. This causes the 3D structures in the background region to mingle with the 3D structures in the foreground region and makes it difficult to create photo-realistic side-view images (\cref{fig:qual_comp_baseline}).
To address this issue, we design our generator $G_\theta$ to separately produce the foreground and background regions to avoid mingled foreground and background structures.
Specifically, our generator $G_\theta$ is composed of two components: an image generator and a background network, inspired by EpiGRAF~\cite{skorokhodov2022epigraf}.
The image generator has two roles: it produces features for the foreground region ({\it i.e.,} the facial region), and produces a final high-resolution image using both foreground and background features.
Meanwhile, the background network produces features for the background region, which are used by the image generator.

For the image generator, we adopt the generator of a state-of-the-art 3D GAN model~\cite{chan2022efficient}.
The image generator forms tri-plane features from the latent code $\mathbf{z}_{fg}$
and the camera parameter $\boldsymbol{\xi}^{+}\in\mathbb{R}^{25}$.
Then, the generator samples 3D positions according to $\boldsymbol{\xi}^{+}$, then obtains features for the sampled positions from the tri-plane.
After that, the foreground feature maps are obtained through a decoder and volume rendering. Then, the feature maps are integrated with the background feature maps according to the transmittance of the foreground to generate a low-resolution feature map. 
Finally, a high-resolution image is obtained from the low-resolution features through a super-resolution module in the image generator. 

For the background network, we adopt the background network of EpiGRAF~\cite{skorokhodov2022epigraf}.
The background network is a multi-layer perceptron (MLP) that takes a latent code $\mathbf{z}_{bg}$ and a 3D position $\mathbf{x}$ as inputs and outputs a feature vector.
To generate background features, we first sample 3D positions according to the camera pose $\boldsymbol{\xi}^{+}$,
and feed them to the background network to obtain feature vectors for the sampled 3D positions.
After aggregating all the background features, we feed them to the image generator.

%%%%%%%%%%%%%%%%%%%%%%%%%%%
% Dual-branched discriminator
%%%%%%%%%%%%%%%%%%%%%%%%%%%
\subsection{Dual-Branched Discriminator}
\label{sec:dual-branched discriminator}

As shown in \cref{fig:main_architecture}, the dual-branched discriminator $D_\phi$ takes an image and a camera pose as inputs.
The input pose can be either positive ($\boldsymbol{\xi}^+$) or negative ($\boldsymbol{\xi}^-$), where a positive pose means that the pose agrees with the input image, while a negative pose means it does not. % agree with the input image.
From the inputs, the discriminator predicts whether the input image is real or fake, and whether the input image agrees with the input camera pose using two output branches.

The discriminator $D_\phi$ comprises four components: a shared block $D_\phi^{\text s}$, a pose encoder $E_\phi$, an image branch $D_\phi^{\text i}$, and a pose branch $D_\phi^{\text p}$.
The shared block extracts features from an input image, which will be used by the image and pose branches,
while the pose encoder $E_\phi$ projects an input camera parameter $\boldsymbol{\xi}$ to an embedding space. 
The image branch $D_\phi^{\text i}$ predicts whether the input image is real or fake using the output of the shared block $D_\phi^{\text s}$. 
The pose branch $D_\phi^{\text p}$ extracts pose features of the input image from the output of shared block $D_\phi^{\text s}$, which are then combined with the features from the pose encoder to discriminate whether the input image agrees with the input camera pose.

%% file: main_tex/4.method_training.tex
\section{Training for a Wider Range of Angles}
\label{sec:training_for_ext_pose}

%%%%%%%%%%%%%%%%%%%%%%%%%%%
% Overview
%%%%%%%%%%%%%%%%%%%%%%%%%%%
In this section, we describe our training strategy including the pose-matching loss and AUPS.

%%%%%%%%%%%%%%%%%%%%%%%%%%%
% 4.1. pose-matching loss
%%%%%%%%%%%%%%%%%%%%%%%%%%%
\subsection{Pose-Matching Loss}
\label{sec:pose-matching loss}

To promote pose consistency between the input pose to the generator and its corresponding synthesized image, the pose-matching loss is computed between a pair of an image and a camera pose.
The pose-matching loss considers both positive and negative pairs of an image and a camera pose to more strongly guide the generator to produce pose-consistent images.
In the case of a positive pair whose image and camera pose are supposed to agree with each other, the pose-matching loss penalizes the generator if the image does not agree with the pose.
On the other hand, in the case of a negative pair whose image and camera pose are supposed to not agree, the pose-matching loss penalizes the generator if the image agrees with the pose.

Formally, we define the pose-matching loss $\mathcal{L}_\textrm{pose}^\textrm{gen}$ for the generator as:
\begin{equation}
\begin{aligned}
\mathcal{L}_{\text{pose}}^\textrm{gen}(\theta) &= \mathcal{L}_{\text{pose}}^{\textrm{gen},+}(\theta) + \mathcal{L}_{\text{pose}}^{\textrm{gen},-}(\theta)\\
&=\mathbb{E}_{\boldsymbol{\xi}^+ \sim p_{\xi}}[h(-(D_\phi^{\text{sp}}(\hat{\mathbf{I}}) \otimes E_\phi(\boldsymbol{\xi}^+)))]\\
&+\mathbb{E}_{\boldsymbol{\xi}^- \sim p_{\xi}}[h(D_\phi^{\text {sp}}(\hat{\mathbf{I}}) \otimes E_\phi(\boldsymbol{\xi}^-))],
\end{aligned}
\end{equation}
where $\otimes$ is an element-wise multiplication, $D_\phi^{\text {sp}}(\cdot) = D_\phi^{\text p}(D_\phi^{\text s}(\cdot))$, $\hat{\mathbf{I}}=G_\theta(\mathbf{z}, \boldsymbol{\xi}^{+})$, and $\mathbf{z} = (\mathbf{z_{\text {fg}}}, \mathbf{z_{\text {bg}}})$.
$h$ is the softplus activation function and $p_{\xi}$ is the pose distribution, whose details will be given in~\cref{sec:aups}
A negative pose $\boldsymbol{\xi}^{-}$ is randomly sampled so as not to be the same as the positive pose $\boldsymbol{\xi}^{+}$.
For a generated image $\hat{\mathbf{I}}$, its positive pose $\boldsymbol{\xi}^{+}$ is the rendering pose used in the generator.

We also define a pose-matching loss to train the discriminator as:
\begin{equation}
\mathcal{L}_{\text{pose}}^\textrm{dis}(\phi) = \mathcal{L}_{\text{pose}}^{\textrm{dis},+}(\phi) + \mathcal{L}_{\text{pose}}^{\textrm{dis},-}(\phi),
\end{equation}
where the terms on the right-hand-side are computed using positive and negative pairs, respectively.
Both $\mathcal{L}_{\text{pose}}^{\textrm{dis},+}(\phi)$ and $\mathcal{L}_{\text{pose}}^{\textrm{dis},-}(\phi)$ are defined using both real and synthesized images for positive and negative pairs.
Specifically, $\mathcal{L}_{\text{pose}}^{\textrm{dis},+}$ is defined as:
\begin{equation}
\begin{aligned}
\mathcal{L}_{\text{pose}}^{\textrm{dis},+}(\phi)&=\mathbb{E}_{(\mathbf{I}, \boldsymbol{\xi}^+) \sim (p_{r},p_{\xi})}[h(-(D_\phi^{\text {sp}}(\mathbf{I}) \otimes E_\phi(\boldsymbol{\xi}^+)))] \\
&+ \mathbb{E}_{\boldsymbol{\xi}^+ \sim p_{\xi}}[h(-(D_\phi^{\text {sp}}(\hat{\mathbf{I}}) \otimes E_\phi(\boldsymbol{\xi}^+)))],
\end{aligned}
\end{equation}
where $p_r$ is the distribution of real images and $\mathbf{I}$ is a real image.
The first and second terms on the right-hand side use real and synthesized pairs as positive pairs, respectively.
For the first term, we sample a real image $\mathbf{I}$ and its corresponding ground-truth pose $\boldsymbol{\xi}^+$ as a positive sample.
The pose-matching loss $\mathcal{L}_{\text{pose}}^{\textrm{dis},-}(\phi)$ for a negative pose is defined as:
\begin{equation}
\begin{aligned}
\mathcal{L}_{\text{pose}}^{\textrm{dis},-}(\phi)&=\mathbb{E}_{\mathbf{I} \sim p_{r},\boldsymbol{\xi}^- \sim p_{\xi}}[h(D_\phi^{\text {sp}}(\mathbf{I}) \otimes E_\phi(\boldsymbol{\xi}^-))] \\
&+\mathbb{E}_{\boldsymbol{\xi}^- \sim p_{\xi}}[h(D_\phi^{\text {sp}}(\hat{\mathbf{I}}) \otimes E_\phi(\boldsymbol{\xi}^-))].
\end{aligned}
\end{equation}

Note that both positive and negative pairs of the pose-matching loss for the discriminator are defined using both real and synthesized images.
Thanks to this, the pose branch of the discriminator is trained to focus only on the pose consistency regardless of whether an image looks real or fake, and subsequently, resulting in the generator being trained to produce pose-consistent images.

%%%%%%%%%%%%%%%%%%%%%%%%%%%
% Final Loss 
%%%%%%%%%%%%%%%%%%%%%%%%%%%

\subsection{Final Loss}
\label{sec:final_loss}
In addition to the pose-matching loss, we adopt other loss terms in our final loss as described in the following.

\noindent{\bf Non-Saturating GAN Loss.}
In the dual-branched discriminator, the image branch $D_\phi^{\text i}$ is optimized by a non-saturating GAN loss to learn the entire target image distribution.
The non-saturating GAN loss for the generator is defined as
\begin{equation}
\begin{aligned}
\mathcal{L}_{\text{adv}}^\textrm{gen}(\theta)&=\mathbb{E}_{\mathbf{z} \sim p_{z},\boldsymbol{\xi}^+ \sim p_{\xi}}[h(-D_\phi^{\text {si}}(G_\theta(\mathbf{z}, \boldsymbol{\xi}^+)))],
\end{aligned}
\end{equation}
where $D_\phi^{\text {si}}(\cdot) = D_\phi^{\text i}(D_\phi^{\text s}(\cdot))$.
The non-saturating GAN loss for the discriminator with $R1$ regularization~\cite{chan2022efficient} is defined as
\begin{equation}
\begin{aligned}
\mathcal{L}_{\text{adv}}^\textrm{dis}(\phi)&=\mathbb{E}_{\mathbf{z} \sim p_{z},\boldsymbol{\xi}^+ \sim p_{\xi}}[h(D_\phi^{\text {si}}(G_\theta(\mathbf{z}, \boldsymbol{\xi}^+)))]\\
&+\mathbb{E}_{\mathbf{I} \sim p_{r}}[h(-D_\phi^{\text {si}}({\mathbf{I}}))
+ \lambda_{R1} |\nabla D_\phi^{\text {si}}({\mathbf{I}})|^{2}],
%+ \lambda |\nabla D_\phi^{\text {si}}({\mathbf{I}})|^{2}],
\end{aligned}
\end{equation}
where $\lambda_{R1}$ is a balancing weight.

% 1) Setting 1 w/o transfer learning
\begin{figure*}[!t]
\includegraphics[width=1\linewidth]{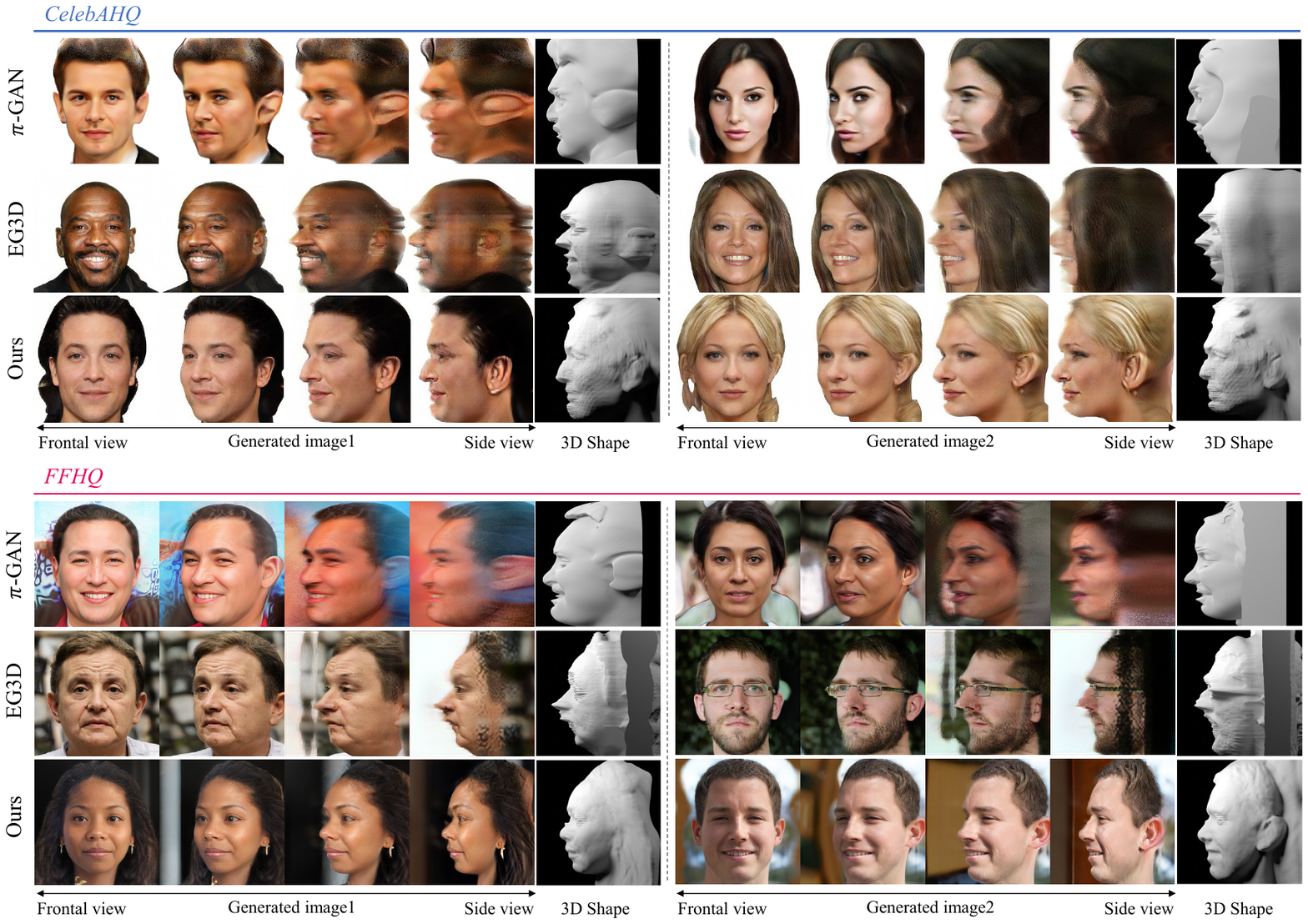}
% \vspace{-0.2cm}
\caption{
Qualitative comparison among $\pi$-GAN~\cite{chan2021pi}, EG3D~\cite{chan2022efficient} and ours.
All the models are trained without transfer learning.
Unlike blurry images and noisy geometry of baselines at the steep pose, our method generates high-quality images and shapes on the target datasets.
(Columns 1-4, 6-9 : the results of a 30-degree rotation from the frontal to the side view. Columns 5, 10 : the side views of the shape obtained using the marching cube.)
}
\label{fig:qual_comp_baseline}
\vspace{-3mm}
\end{figure*}

% 2) Setting 1 w/o transfer learning
\begin{figure*}[!t]
\includegraphics[width=1\linewidth]{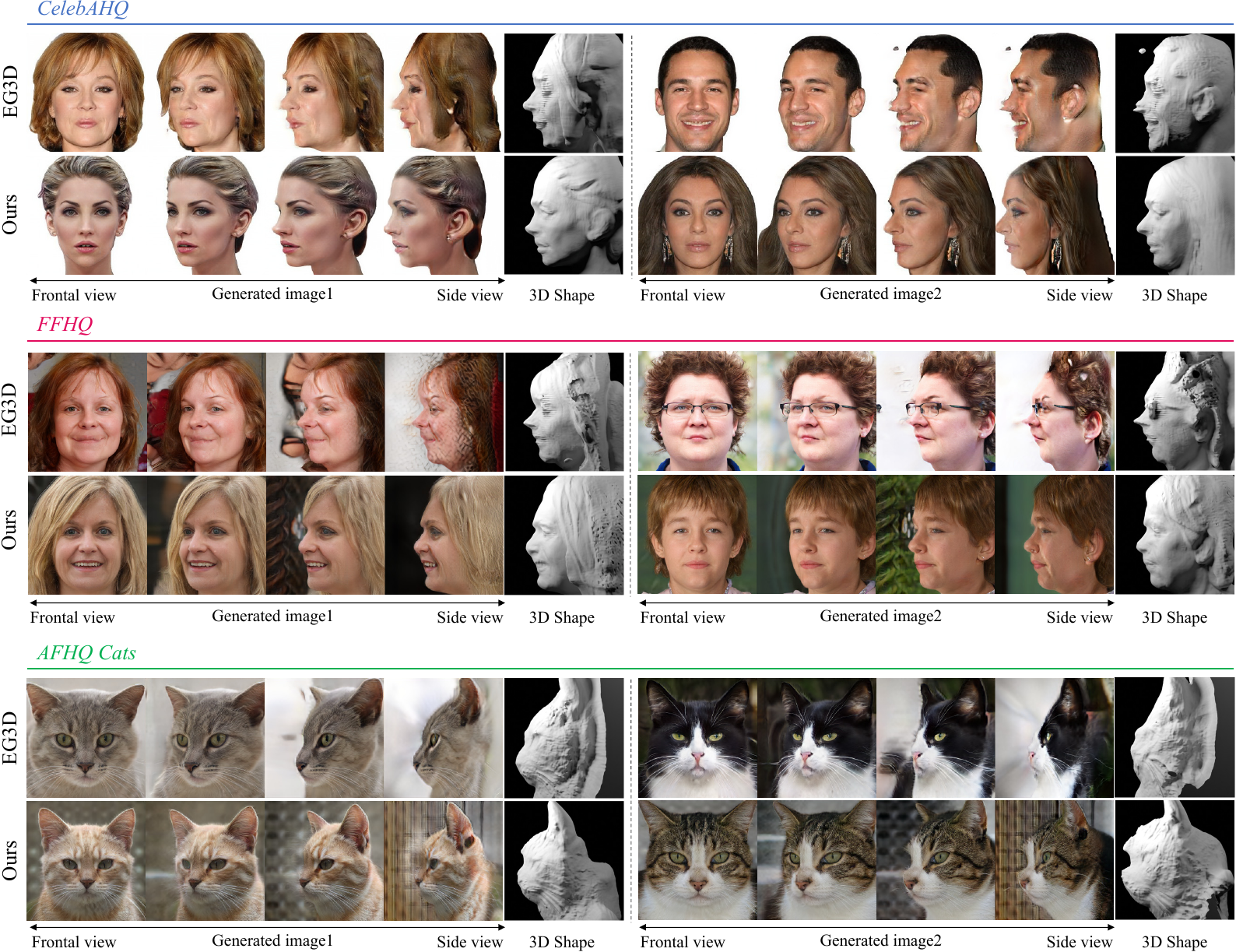}
% \vspace{-0.2cm}
\caption{
Qualitative comparison between EG3D~\cite{chan2022efficient} and 
ours.
Both models are trained with transfer learning.
Unlike unnatural images and geometry of the baseline at the steep pose, our method generates high-quality images and shapes on the target datasets.
(Columns 1-4, 6-9 : the results of a 30-degree rotation from the frontal to the side view. Columns 5, 10 : the side views of the shape obtained using the marching cube.)
% \wj{CelebAHQ-EG3D 2개, FFHQ-EG3D 2개, Cats-EG3D 2개}
}
\label{fig:qual_comp_baseline_transfer}
\vspace{-3mm}
\end{figure*}

\noindent{\bf Identity Regularization.}
%%%%%%%%%%%%%%%%%%%%%%%%%%%%%%%%%%%%%%%%%%%%%%%%%%%%%%%%
To encourage the generator to create semantically various images, we train the generator with an additional identity regularization term $\mathcal{L}_{\text{id}}=\lambda_{\text{z}}\mathcal{L}_{\text{z}}+\lambda_{\text{c}}\mathcal{L}_{\text{c}}$,
where $\mathcal{L}_{\text{z}}$ is a loss term to promote images with diverse identities,
and $\mathcal{L}_{\text{c}}$ is a term to prevent the identity of a generated image from being affected by the camera parameter $\boldsymbol{\xi}$.
$\lambda_{\text{z}}$ and $\lambda_{\text{c}}$ are balancing weights.
$\mathcal{L}_{\text{z}}$ is defined as
\begin{equation}
\begin{aligned}
&\mathcal{L}_{\text{z}}(\theta)=\mathbb{E}_{\mathbf{z}_{\text 1},\mathbf{z}_{\text 2} \sim p_{z},\boldsymbol{\xi}^{+} \sim p_{\xi}}[\langle E_{id}(\hat{\mathbf{I}}_{1}), E_{id}(\hat{\mathbf{I}}_{2})\rangle],\\
\end{aligned}
\end{equation}
where $\hat{\mathbf{I}}_{1}=G_\theta(\mathbf{z}_{\text 1}, \boldsymbol{\xi}^{+})$, $\hat{\mathbf{I}}_{2}=G_\theta(\mathbf{z}_{\text 2}, \boldsymbol{\xi}^{+})$, and $E_{\text {id}}$ is a face identity network~\cite{deng2019arcface}.
$\langle \cdot,\cdot \rangle$ calculates the cosine similarity. 
$\mathcal{L}_{\text{c}}$ is defined as:
\begin{equation}
\begin{aligned}
&\mathcal{L}_{\text{c}}(\theta)=\mathbb{E}_{\mathbf{z} \sim p_{z},\boldsymbol{\xi}^{+}_{\text 1},\boldsymbol{\xi}^{+}_{\text 2} \sim p_{\xi}}\left[\frac{1 - \langle E_{\text {id}}(\hat{\mathbf{I}}_{1}), E_{id}(\hat{\mathbf{I}}_{2})\rangle}{\parallel \hat{\mathbf{I}}_{1} - \hat{\mathbf{I}}_{2} \parallel_{1}}\right],\\
\end{aligned}
\end{equation}
where $\hat{\mathbf{I}}_{1}=G_\theta(\mathbf{z}, \boldsymbol{\xi}^{+}_{\text 1})$, and $\hat{\mathbf{I}}_{2}=G_\theta(\mathbf{z}, \boldsymbol{\xi}^{+}_{\text 2})$.
The identity regularization $\mathcal{L}_{\text{id}}$ helps the generator faithfully learn semantic information from the dataset, enabling image synthesis with high fidelity (\cref{sec:ablations}).
\\\\
\noindent{\bf Final Loss.}
The final losses for training the generator and the discriminator are then defined as:
\begin{equation}
\begin{aligned}
&\mathcal{L}_{\text{total}}^\textrm{gen}=\mathcal{L}_{\text{adv}}^\textrm{gen}+\lambda_{\text{pose}}\mathcal{L}^\textrm{gen}_{\text{pose}}+\mathcal{L}_{\text{id}}+\lambda_{\text{d}}\mathcal{L}_{\text{d}},~~~\textrm{and}\\
&\mathcal{L}_{\text{total}}^\textrm{dis}=\mathcal{L}_{\text{adv}}^\textrm{dis}+\lambda_{\text{pose}}\mathcal{L}^\textrm{dis}_{\text{pose}},
\end{aligned}
\end{equation}
where 
$\mathcal{L}_{\text{d}}$ is an additional $L^1$-based density regularization term~\cite{sun2022ide}. $\lambda_{\text{pose}}$ and $\lambda_{\text{d}}$ are weights to balance the terms.
More details on the losses can be found in~\cref{sec:experiment}.

\subsection{Additional Uniform Pose Sampling}
\label{sec:aups}

As previous methods mostly focus on learning frontal-view images in their training because of the pose-imbalanced dataset, they lack opportunities for learning side-view images, resulting in degenerate side-view image quality.
To increase the opportunities of learning side-view images in pose-imbalanced datasets, our AUPS strategy samples camera poses for rendering fake images from the training dataset like EG3D~\cite{chan2022efficient} and additionally sample poses from a uniform distribution in training.
Specifically, for computing the non-saturating GAN loss with the image branch of the discriminator, we use camera poses sampled from the training dataset and the uniform distribution together.
For computing the pose-matching loss and the identity regularization with the pose branch of the dual-branched discriminator, on the other hand, we simply use camera poses sampled only from the training dataset like EG3D as we found that the pose branch can already be effectively trained without AUPS.

While sampling camera poses solely from the uniform distribution may seem straightforward to increase the learning opportunities at steep angles, it can lead to a significant discrepancy between the real and fake image distributions, which may harm the training process.
To mitigate this, we use both pose distribution of the training dataset and uniform distribution together to decrease the distribution discrepancy while increasing learning opportunities for steep angles.
More details on the AUPS can be found in Sec.~B.3.

%% file: main_tex/5.experiments.tex
\section{Experiments}
\label{sec:experiment}
\noindent{\bf Implementation Details.}
\label{sec:training_detail}
Most of the experimental setups and preprocessing methods are the same as those of EG3D~\cite{chan2022efficient} except for the following. We set the dimensions of the background latent vector $\mathbf{z}_{bg}$ to 512. The final image resolution of our model is $256\times 256$ and the neural rendering resolution is fixed as $64\times 64$. The neural rendering result is bilinearly upsampled to $128\times128$ and fed to the super-resolution module in the image generator. 
The batch size is set to 64 in all the experiments. 
The balancing weights for the loss terms are set as follows:
$\lambda_{\text{pose}} = 1$, $\lambda_{\text{z}} = 0.5$, $\lambda_{\text{c}} = 0.25$, $\lambda_{\text{d}} = 0.25$ and $\lambda_{R1}$ = 1.

\noindent{\bf Datasets.}
\label{sec:datasets}
We validate our method on real-world human face datasets (CelebAHQ~\cite{karras2017progressive} and FFHQ~\cite{karras2019style}) and a real-world cat face dataset (AFHQ Cats~\cite{choi2020stargan}). To show results both with and without background regions, we remove the background regions of the CelebAHQ dataset using the ground-truth segmentation masks, but keep the background regions of the FFHQ dataset in our experiments. We obtain the ground-truth poses of real images using pre-trained camera pose estimation models~\cite{feng2021learning,cats_campose}.

\noindent{\bf Transfer learning.}
As used in previous 3D GANs for compensating for the small dataset size, we optionally adopt transfer learning to improve the quality of side-view image synthesis~\cite{chan2022efficient,gu2021stylenerf}. 
To be specific, we pre-train a generator with a pose-balanced synthetic dataset and fine-tune it with a pose-imbalanced in-the-wild dataset to compensate for insufficient knowledge for side-view images in in-the-wild datasets.
Specifically, we use the FaceSynthetics dataset~\cite{wood2021fake} for pre-training. 
We also remove the background regions of the FaceSynthetics dataset with the ground-truth segmentation masks to accurately learn 3D geometries.
We use the training strategy of EG3D~\cite{chan2022efficient} to pre-train models.

%%%%%%%%%%%%%%%%%%%%%%%%%%%
% Experimental Results
%%%%%%%%%%%%%%%%%%%%%%%%%%%
% \subsection{Experimental Results}
\subsection{Comparison}
\label{sec:comparison}
We first conduct qualitative and quantitative comparisons of \MethodName{} and previous 3D GANs ($\pi$-GAN~\cite{chan2021pi} and EG3D~\cite{chan2022efficient}) on different datasets (CelebAHQ~\cite{karras2017progressive}, FFHQ~\cite{karras2019style} and AFHQ Cats~\cite{choi2020stargan}) both with and without transfer learning.

\begin{table}[!ht]
\resizebox{\columnwidth}{!}{
\begin{tabular}{c|ccc|cc}
\hline
& \multicolumn{3}{c|}{FID$\downarrow$}& \multicolumn{2}{c}{Depth error$\downarrow$}                      \\ \hline
Method \textbackslash Dataset & \multicolumn{1}{c}{CelebAHQ}        & \multicolumn{1}{c}{FFHQ}            & AFHQ(Cats)      & \multicolumn{1}{c}{CelebAHQ}       & FFHQ           \\ \hline
$\pi$-GAN  & \multicolumn{1}{c}{80.372}          & \multicolumn{1}{c}{120.991}         & -         & \multicolumn{1}{c}{2.438}          & 1.365          \\ 
EG3D    & \multicolumn{1}{c}{40.760}          & \multicolumn{1}{c}{35.348}          & -          & \multicolumn{1}{c}{0.760}          & 0.921          \\ 
Ours    & \multicolumn{1}{c}{\textbf{37.417}} & \multicolumn{1}{c}{\textbf{22.174}} & - & \multicolumn{1}{c}{\textbf{0.580}} & \textbf{0.649} \\ \hline\hline
EG3D+transfer learning    & \multicolumn{1}{c}{28.912}          & \multicolumn{1}{c}{26.627}          & 15.639          & \multicolumn{1}{c}{0.606}          & 0.864          \\ 
Ours+transfer learning    & \multicolumn{1}{c}{\textbf{22.219}} & \multicolumn{1}{c}{\textbf{24.571}} & \textbf{10.134} & \multicolumn{1}{c}{\textbf{0.549}} & \textbf{0.657} \\ \hline
\end{tabular}
}
\caption{Quantitative comparison of the image and shape quality with baselines. 
}
\label{tab:quant_baseline}
\vspace{-5mm}
\end{table}

\noindent{\bf Qualitative Comparison.}
\cref{fig:qual_comp_baseline} shows a qualitative comparison between our method and the previous 3D GANs.
In this comparison, all the models are trained from scratch without transfer learning.
The AFHQ Cats dataset~\cite{choi2020stargan} is not included in this comparison as the dataset is too small to train a generator without transfer learning.
For all the real-world human face datasets, $\pi$-GAN and EG3D generate blurry images for steep angles compared to realistic frontal images.
In contrast, \MethodName{} robustly generates high-quality images irrespective of camera pose. 
\cref{fig:qual_comp_baseline_transfer} shows another qualitative comparison where we adopt transfer learning.
For all the datasets, EG3D generates unnatural images for steep angles compared to realistic frontal images.
On the other hand, \MethodName{} robustly generates high-quality images irrespective of camera pose. 
These results indicate that our method is effective in learning to synthesize high-quality images at all camera poses in both cases with and without transfer learning.
Additional results are in Sec.~D.

\begin{figure}[!t]
\includegraphics[width=1\linewidth]{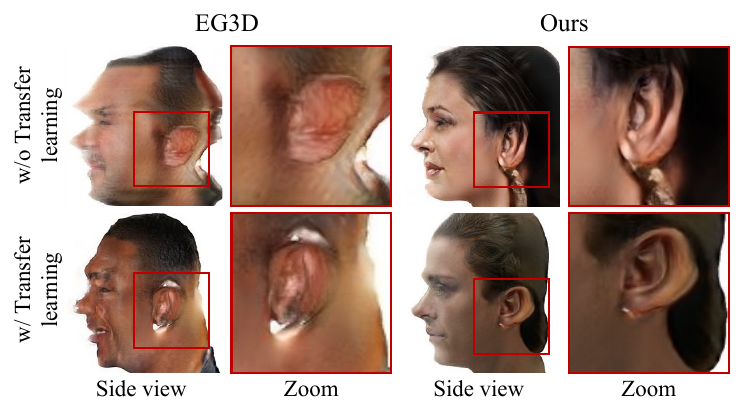}
\vspace{-0.5cm}
\caption{Visual comparison of side-view images on CelebAHQ~\cite{karras2017progressive} on the setting with or without transfer learning.
Without transfer learning, our proposed method outperform the baseline (EG3D~\cite{chan2022efficient}), which shows noisy facial boundaries.
With transfer learning, our method also outperforms the baseline, which generates holes.
}
\label{fig:ablation_vis}
\vspace{-3mm}
\end{figure}

\cref{fig:ablation_vis} shows zoomed-in patches of side-view images of \MethodName{} and EG3D~\cite{chan2022efficient} to compare the quality of synthesized details.
As the figure shows, \MethodName{} produces more realistic details for side-view images with much less artifacts than EG3D regardless of transfer learning.
In addition, the figure also shows that transfer learning helps both models generate clearer images as it provides additional information on side views of human faces.
Nevertheless, the result of EG3D with transfer learning still suffers from severe artifacts such as holes due to its pose-sensitive training process.

%%%%%%%%%%%%%%%%%%%%%%%%%%%
% Quantitative comparison
%%%%%%%%%%%%%%%%%%%%%%%%%%%
\noindent{\bf Quantitative Comparison.}
We conduct a quantitative evaluation on the image and shape quality.
To evaluate the pose-irrespective performance of the models, we generate images and shapes at randomly sampled camera poses from a uniform distribution.
Refer to Sec.~C.5. for more details regarding the pose sampling strategy used in this experiment.
\cref{tab:quant_baseline} shows the quantitative comparison. As the table shows, in both cases with and without transfer learning, \MethodName{} outperforms all the other baselines in terms of image quality based on FID~\cite{heusel2017gans}
thanks to our effective training method.

% 2) ICCV version

Due to the absence of 3D geometries corresponding to synthesized images, we evaluate the shape quality with pseudo-ground-truth shapes, which are estimated from synthesized images using an off-the-shelf 3D reconstruction model~\cite{feng2021learning}, as done in EG3D~\cite{chan2022efficient}.
We measure depth error by calculating MSE between generated depth from our model and rendered depth from the estimated geometry. \cref{tab:quant_baseline} shows that \MethodName{} achieves the best depth accuracy than the other baselines for the shape quality both with and without transfer learning.
This remarkable improvement can also be shown in \cref{fig:qual_comp_baseline} and \cref{fig:qual_comp_baseline_transfer}, where generated shapes from \MethodName{} show high-fidelity 3D geometries compared to those of the other methods.

\begin{table}[t!]
\centering
\resizebox{\columnwidth}{!}{%
\begin{tabular}{c|c|c|c|c}
\hline
 & AUPS & \begin{tabular}[c]{@{}c@{}}Dual-branched discriminator\\ \& Pose-matching loss\end{tabular} & \begin{tabular}[c]{@{}c@{}} Identity\\regularization\\ ($\mathcal{L}_{\text {id}}$)\end{tabular} & FID$\downarrow$ \\ \hline
\multirow{2}{*}{EG3D} &  &  &  & 28.912 \\ \cline{2-5} 
 & \checkmark &  &  & 30.553 \\ \hline
\multirow{2}{*}{\begin{tabular}[c]{@{}c@{}}SideGAN\\ (Ours)\end{tabular}} & \checkmark & \checkmark &  & 23.106 \\ \cline{2-5} 
 & \checkmark & \checkmark & \checkmark & \textbf{22.219} \\ \hline
\end{tabular}%
}
\vspace{-0.3cm}
\caption{Ablation study for key components of the proposed method on CelebAHQ~\cite{karras2017progressive}. }
\label{table:abl_key}
\end{table}

\begin{table}[t!]
\resizebox{\columnwidth}{!}{
\begin{tabular}{c|c|c}
\hline
                            & FID$\downarrow$ & Depth error$\downarrow$ \\ \hline
Ours w/ pose-regression loss & 30.069             & 0.624          \\ \hline
Ours w/ pose-matching loss   & \textbf{22.219}    & \textbf{0.549} \\ \hline
\end{tabular}
}
\vspace{-0.3cm}
\caption{Comparison between the pose-matching loss and the pose-regression loss~\cite{deng2022gram} on CelebAHQ~\cite{karras2017progressive}.
}
\label{table:abl_reg}
% \vspace{-3mm}
\end{table}

\begin{figure}[t!]
% \vspace{-2mm}
\centering
\includegraphics[width=1.0\linewidth]{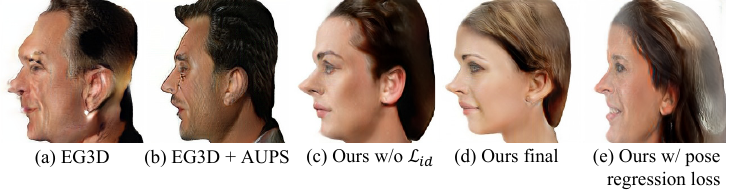}
\vspace{-0.5cm}
\caption{\small{Additional visual results of ablation study. (b) While AUPS helps improve side-view image quality, artifacts still remain. (d) $\mathcal{L}_{\text {id}}$ results in a slightly clearer side-view image than (c). (e) Instead of our pose-matching loss, our model with the pose-regression loss leads to a flattened shape.
% seed: 0-50 사이
}}
\label{fig:ablation_extra_sup}
\vspace{-3mm}
\end{figure}

%%%%%%%%%%%%%%%%%%%%%%%%%%%
% Ablation studies
%%%%%%%%%%%%%%%%%%%%%%%%%%%
\subsection{Ablation Studies}
\label{sec:ablations}

We conduct ablation studies to evaluate the benefits of three components in our framework:
1) the dual-branched discriminator (\cref{sec:dual-branched discriminator}), 2) the pose-matching loss (\cref{sec:pose-matching loss}), 3) AUPS (\cref{sec:aups}),
and 4) the identity regularization $\mathcal{L}_{\text{id}}$ (\cref{sec:final_loss}).
The ablation studies are conducted using the CelebAHQ dataset~\cite{karras2017progressive}.

\cref{table:abl_key} and \cref{fig:ablation_extra_sup} report the ablation study result.
In \cref{table:abl_key}, the pose-matching loss and the dual-branched discriminator are applied together since supervision is needed for the pose branch of the discriminator.
With AUPS, the image quality of side-view is improved in EG3D~\cite{chan2022efficient} (\cref{fig:ablation_extra_sup} (b)) since AUPS increases the learning opportunities at steep angles.
However, the side-view images still have artifacts and the FID of EG3D deteriorates. This is because EG3D learns the real/fake distribution in a pose-wise manner through a pose-conditional GAN loss, which is unstable under the misalignment between two distributions, caused by AUPS as mentioned in \cref{sec:aups}. 
Unlike EG3D, our framework with AUPS improves the FID as each component is added, proving the benefit of each component. This is because SideGAN's GAN loss is more robust to the mismatch of the pose distribution than EG3D and our model learns photo-realism and pose-consistency separately through the dual-branched discriminator and the pose-matching loss.

To evaluate the effectiveness of the pose-matching loss in learning side-view images and 3D geometries, we compare our pose-matching loss with the pose-regression loss of GRAM~\cite{deng2022gram} both quantitatively and qualitatively.
As shown in \cref{table:abl_reg}, our pose-matching loss results in a significantly lower FID score and depth error, owing to the fact that our binary-classification-based pose-matching loss allows for easier training.
We also provide visual comparison in \cref{fig:ablation_extra_sup}. Compared to our model trained with the pose-matching loss (d), the model trained with the pose-regression loss (e) produces a flattened shape, demonstrating the advantages of the pose-matching loss.

%%%%%%%%%%%%%%%%%%%%%%%%%%%
% Evaluation at steep and extrapolated angles
%%%%%%%%%%%%%%%%%%%%%%%%%%%
\subsection{Effects on the Steep and Extrapolated Angles}
Finally, we conduct a more detailed quantitative analysis of \MethodName{} for different camera poses by measuring the FID scores of synthesized images for frontal, steep, and extrapolated angles.
Measuring FID scores requires a sufficient amount of ground-truth images for each camera pose, which is not the case for the in-the-wild datasets.
Thus, we conduct our analysis using the FaceSynthetic dataset~\cite{wood2021fake}, which is pose-balanced and provides a larger number of images for a wider range of camera poses compared to the in-the-wild datasets. 
Specifically, we first construct a pose-imbalanced training dataset from FaceSynthetics by randomly sampling images within the pose range from $-50^\circ$~to $50^\circ$~to have a Gaussian pose distribution like in-the-wild datasets.
Then, we train our model with the dataset, and evaluate its FID scores for different camera poses using the original FaceSynthetics dataset.
For comparison, we also evaluate the FID scores of EG3D~\cite{chan2022efficient}.
In this experiment, we did not apply transfer learning.

\begin{figure}[!t]
\centering
\includegraphics[width=0.85\linewidth]{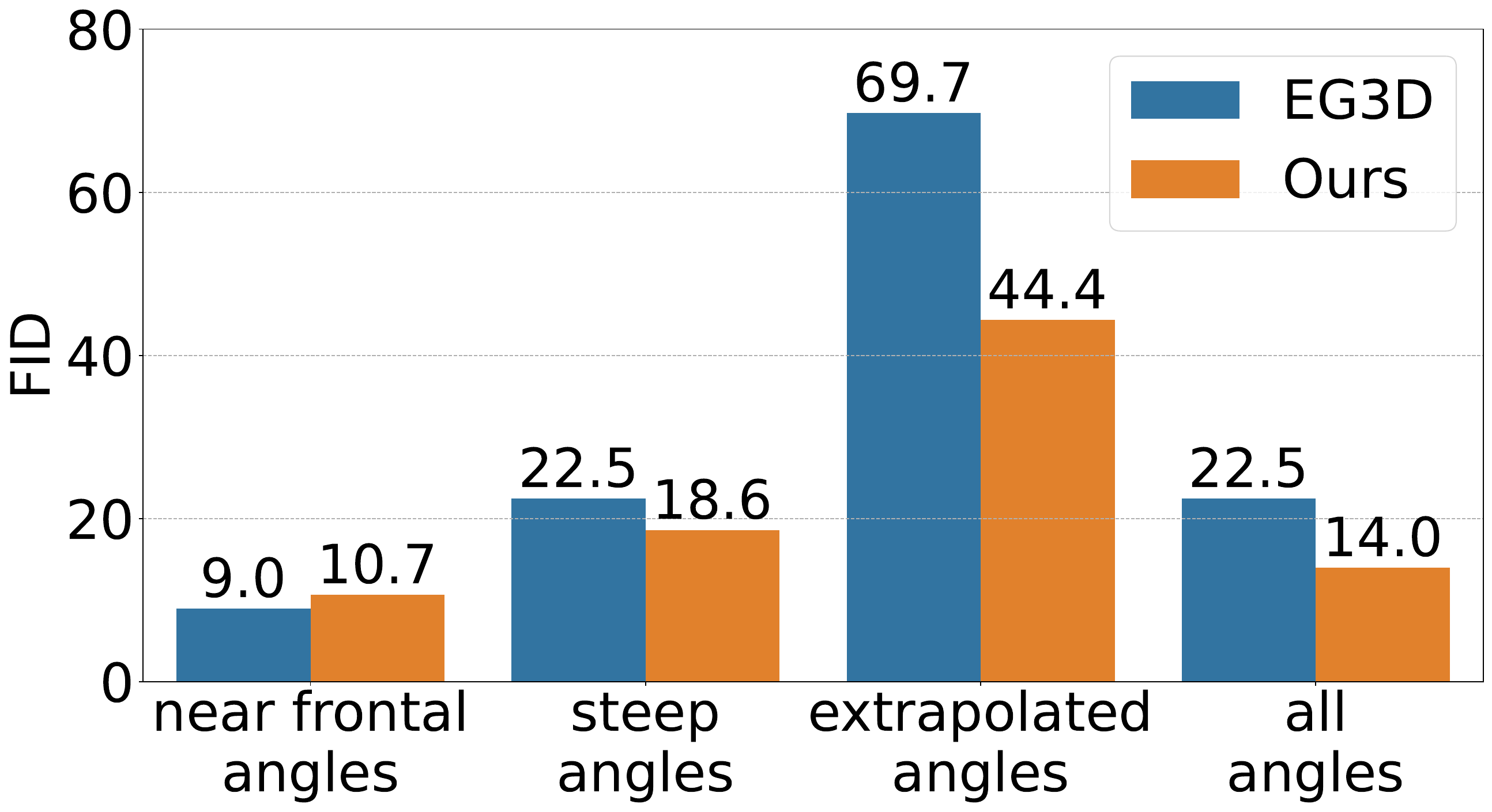}
 \vspace{-0.3cm}
\caption{Comparison on image quality (FID$\downarrow$) with regard to the range of the camera angles. We limit the FaceSynthetic dataset~\cite{wood2021fake} not to have any images within the range of extrapolated angles. \MethodName{} outperforms EG3D~\cite{chan2022efficient} in image quality except for the range of the frontal view, which shows even competitive result. All angles are from -90 to 90 degrees based on the frontal view.
}
\label{fig:steep_extrapol}
\vspace{-2mm}
\end{figure}

\cref{fig:steep_extrapol} shows the evaluation result for different camera poses.
In the figure, the near-frontal angles are $(-30^\circ, 30^\circ)$, and the steep angles are $(-50^\circ, -30^\circ)\bigcup(30^\circ, 50^\circ)$.
The extrapolated angles indicate angles smaller or larger than $-50^\circ$ and $50^\circ$, which are outside the training distribution.
As shown in the figure, our model performs comparably to EG3D at near-frontal angles, and as the angle gets larger, our model performs significantly better than EG3D, proving the effectiveness of our approach.

%% file: main_tex/6.conclusion.tex
\section{Conclusion}
\label{sec:conclusion}
In this paper, we proposed \MethodName{}, a novel 3D GAN training method to generate high-quality images irrespective of the camera pose.
Our method is based on the key idea that decomposes the originally challenging problem into two easier subproblems, each of which promotes pose consistency and photo-realism, respectively.
Based on this, we propose a novel dual-branched discriminator and a pose-matching loss.
We also presented AUPS to increase the learning opportunities for improving the synthesis quality at a side viewpoint.

Our experimental results show that our method can synthesize photo-realistic images irrespective of the camera pose on human and animal face datasets. 
Especially, even only with pose-imbalanced in-the-wild datasets, our model can generate details of side-view images such as ears, unlike blurry images from the baselines.

Our method is not free from limitations.
For animal faces, we found that black spot-like artifacts appear behind the ear, which might be due to the lack of knowledge about the back of the ear since we conduct transfer learning from synthetic human face to animal face.
Also, despite the background network, the background region is sometimes not clearly separated.
However, we expect that a more advanced background separation scheme such as~\cite{shin2023ballgan} would be able to resolve this.